\documentclass[10pt,twocolumn,letterpaper]{article}

\usepackage{iccv}
\usepackage{times}
\usepackage{epsfig}
\usepackage{graphicx}
\usepackage{amsmath}
\usepackage{amssymb}
\usepackage{bbm}
\usepackage{algorithm}
\usepackage{algorithmic}

\usepackage{multirow}
\usepackage{color}
\usepackage{float}
\usepackage{caption}
\usepackage{subcaption}
\usepackage[toc,page]{appendix}
\usepackage[pagebackref=true,breaklinks=true,letterpaper=true,colorlinks,bookmarks=false]{hyperref}

\iccvfinalcopy % *** Uncomment this line for the final submission

\def\iccvPaperID{6980} % *** Enter the ICCV Paper ID here

\ificcvfinal\fi

\newcommand{\figref}[1]{Fig.~\ref{#1}}
\newcommand{\tabref}[1]{Tab.~\ref{#1}}
\newcommand{\secref}[1]{Sec.~\ref{#1}}
\newcommand{\algref}[1]{Algorithm~\ref{#1}}
\newcommand{\equref}[1]{Equ.~(\ref{#1})}

\begin{document}

%%%%%%%%% TITLE
\title{Meta Gradient Adversarial Attack}

\author{Zheng Yuan$^{1,2}$, ~Jie Zhang$^{1,2}$, ~Yunpei Jia$^{1,2}$, ~Chuanqi Tan$^{3}$, ~Tao Xue$^{3}$, ~Shiguang Shan$^{1,2}$\\
$^{1}$Institute of Computing Technology, Chinese Academy of Sciences\\ $^{2}$University of Chinese Academy of Sciences \quad $^{3}$ Tencent\\ 
{\tt\small \{zheng.yuan, yunpei.jia\}@vipl.ict.ac.cn}; {\tt\small\{zhangjie,sgshan\}@ict.ac.cn};\\ {\tt\small\{jamestan,emmaxue\}@tencent.com}
% For a paper whose authors are all at the same institution,
% omit the following lines up until the closing ``}''.
% Additional authors and addresses can be added with ``\and'',
% just like the second author.
% To save space, use either the email address or home page, not both
% \and
% author2\\
% Institution2\\
% First line of institution2 address\\
% {\tt\small secondauthor@i2.org}
}

\maketitle
\thispagestyle{empty}

%%%%%%%%% ABSTRACT
\begin{abstract}
   In recent years, research on adversarial attacks has become a hot spot. Although current literature on the transfer-based adversarial attack has achieved promising results for improving the transferability to unseen black-box models, it still leaves a long way to go. Inspired by the idea of meta-learning, this paper proposes a novel architecture called Meta Gradient Adversarial Attack (MGAA), which is plug-and-play and can be integrated with any existing gradient-based attack method for improving the cross-model transferability. Specifically, we randomly sample multiple models from a model zoo to compose different tasks and iteratively simulate a white-box attack and a black-box attack in each task. By narrowing the gap between the gradient directions in white-box and black-box attacks, the transferability of adversarial examples on the black-box setting can be improved. Extensive experiments on the CIFAR10 and ImageNet datasets show that our architecture outperforms the state-of-the-art methods for both black-box and white-box attack settings.
\end{abstract}

%%%%%%%%% BODY TEXT
% \setlength{\abovedisplayskip}{2pt} % 对公式上下间隔有用
% \setlength{\belowdisplayskip}{2pt}
% \setlength{\intextsep}{5pt} % Vertical space above & below [h] floats
% \setlength{\textfloatsep}{5pt} % Vertical space below (above) [t] ([b]) floats
% \setlength{\parskip}{0mm}

% \setlength{\abovecaptionskip}{5pt}
% \setlength{\belowcaptionskip}{5pt}

\section{Introduction}
With the rapid development of neural networks in recent years, the reliability of neural networks has gradually attracted more and more attention. The neural networks are exceedingly sensitive to adversarial examples, \ie, the imperceptible perturbation on the input can easily fool the model, leading to unexpected mistakes. For example, when employing face recognition technology for payment, a slight perturbation on the face image may trick the face recognition model into recognizing as someone else. Since attack and defense are two complementary aspects, the researches on adversarial attacks can ultimately improve the robustness of the model, thereby making the model more reliable.

In recent years, many methods have been proposed to improve the success rate of attacks against white-box models, such as FGSM~\cite{goodfellow2014explaining}, C\&W~\cite{carlini2017towards}, PGD~\cite{madry2017towards}, BIM~\cite{kurakin2016adversarial}, DeepFool~\cite{moosavi2016deepfool}, \etc. Based on the access to model parameters, these methods can make the model misclassify the input images by only adding human-imperceptible perturbations, which is named as the white-box attack. However, in reality, a more practical scenario is that the attacker cannot obtain any information of the target model, that is, the black-box attack. Therefore, some methods turn to utilize the transferability of adversarial examples to conduct black-box attacks, such as MIM~\cite{dong2018boosting}, DIM~\cite{xie2019improving}, TIM~\cite{dong2019evading}, NI-SI~\cite{lin2019nesterov}, \etc. Although most of these methods have achieved promising results under the scenario of black-box attacks, the transferability of adversarial examples is still limited due to the discrepancy between the white-box models and unseen black-box models.
\begin{figure*}[htbp]
   \centering
   \includegraphics[width=0.9\textwidth]{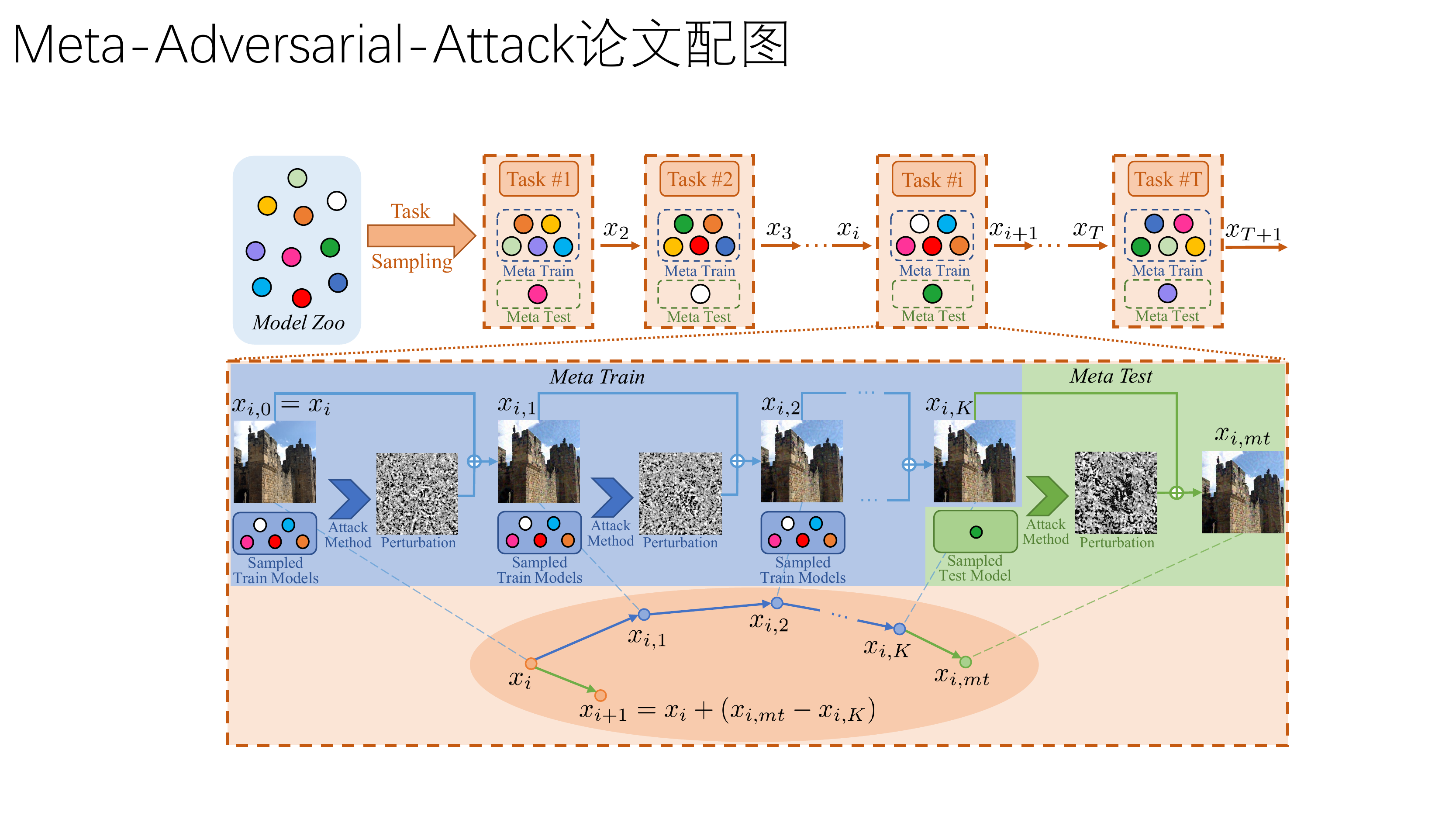}
   \vspace{-3mm}
   \caption{Overview of our Meta Gradient Adversarial Attack (MGAA). MGAA consists of multiple ($T$ in the figure) iterations. In each iteration, $n+1$ models are randomly sampled from the model zoo to compose a meta-task. Each task is divided into two steps: the meta-train step and the meta-test step. In the meta-train step, an ensemble of the $n$ sampled models is utilized to conduct the gradient-based attack to generate perturbations, which can be repeated by $K$ times. The meta-test step uses the adversarial example $x_{i,K}$ obtained from the meta-train step as the basis, and utilizes the last sampled model to generate perturbations for adversarial attacks. Finally, the perturbation generated in the meta-test step $x_{i,mt} - x_{i,K}$ is added to $x_i$, the final adversarial example after the $i$-th task.}
   \label{fig:architecture}
   \vspace{-6mm}
\end{figure*}

Inspired by the philosophy of meta-learning, we propose a novel architecture named Meta Gradient Adversarial Attack (MGAA), which is plug-and-play and can be incorporated with any gradient-based adversarial attack method. The main idea of MGAA is to generate the adversarial examples by iteratively simulating white-box and black-box attacks to improve the transferability. Specifically, as shown in \figref{fig:architecture}, in each iteration, multiple models are randomly sampled from the model zoo to compose a task, which can be divided into the meta-train and the meta-test step. The meta-train step first uses an ensemble of multiple models to simulate white-box attacks to obtain temporary adversarial examples, which are then used as a basis by the meta-test step to obtain the perturbation by simulating a black-box attack scenario. Finally, the perturbation obtained during the meta-test step is added to the adversarial examples generated in the previous iteration. In Section~\ref{sec:analysis}, more theoretical analyses demonstrate that our proposed architecture can gradually narrow the gap of gradient directions between white-box attack and black-box attack settings, thus improving the transferability of generated adversarial examples. Different from vanilla meta-learning methods, which enhance the generalization through model training, our proposed MGAA directly utilizes the gradient information to improve the transferability of the adversarial examples without the need of training an extra model.

% Different from the existing methods, we resort to meta-learning to improve the transferability of adversarial examples under the black-box attack settings. We propose a novel architecture called Meta Gradient Adversarial Attack (MGAA), which generates adversarial examples by iteratively simulating white-box and black-box attacks. Specifically, as shown in \figref{fig:architecture}, each iteration randomly samples multiple models from the model zoo to compose a task, which can be divided into the meta-train and the meta-test steps. The meta-train step firstly uses multiple models to simulate white-box attacks to obtain temporary adversarial examples. And then, the meta-test step utilizes another model to obtain the perturbation by simulating black-box attacks on the basis of temporary adversarial examples. Finally, the perturbation obtained during the meta-test step is added to the adversarial examples generated in the previous iteration. The MGAA architecture is plug-and-play and can be incorporated with any gradient-based adversarial attack method. In Section~\ref{sec:analysis}, more analyses are provided to demonstrate that our proposed architecture can gradually narrow the gap of gradient directions between white-box attacks and black-box attacks settings through theoretical analysis, result in improving the transferability of generated adversarial examples.

Extensive experiments on the CIFAR10~\cite{krizhevsky2009learning} and ImageNet~\cite{russakovsky2015imagenet} dataset demonstrate that our proposed architecture significantly improves the success rates of both white-box and black-box attacks. Especially, by integrating TI-DIM~\cite{dong2019evading} method into our proposed architecture, the average attack success rate under the targeted attack setting against 10 white-box models increases by 27.67\%, and the attack success rate against 6 black-box models increases by 28.52\% on ImageNet, which clearly shows the superiority of our method.

The main contributions of this paper are as follows:

   1. We propose the Meta Gradient Adversarial Attack architecture inspired by the philosophy of meta-learning, which iteratively simulates the white-box and the black-box attack to narrow the gap of the gradient directions when generating adversarial examples, thereby further improving the transferability of adversarial examples.
   
   2. The proposed architecture can be combined with any existing gradient-based attack method in a plug-and-play mode. 
   
   3. Extensive experiments show that the proposed architecture can significantly improve the attack success rates under both white-box and black-box settings.
%------------------------------------------------------------------------
\section{Related Work}
In this section, we will give a brief introduction to the related works, \ie, the adversarial attack, the adversarial defense, and the meta-learning.
% \subsection{Adversarial Example}
% The task of adversarial examples is first proposed in \cite{szegedy2013intriguing}. The work of adversarial examples is mainly divided into two parts: adversarial attack and adversarial defense. The method of adversarial attack adds a slight perturbation to the image, and the target model is expected to misclassify the disturbed image. While the method of adversarial defense is to improve the robustness of the model so that the perturbed adversarial examples can still be correctly classified.
\subsection{Adversarial Attack}
The task of adversarial attack is generally classified into four categories according to the amount of target model information we can access: white-box attack, score-based black-box attack, decision-based black-box attack, and transfer-based black-box attack.

\textbf{White-box Attack.}
The white-box attack can obtain all the information of the target model, including model parameters, model structure, gradient, \etc.
FGSM~\cite{goodfellow2014explaining} utilizes gradient information to update the adversarial example in one step along the direction of maximum classification loss. This method is extended by BIM~\cite{kurakin2016adversarial} to propose an iterative method to generate adversarial examples through multi-step updates. PGD~\cite{madry2017towards} is similar to BIM, except that it randomly selects an initial point in the neighborhood of the benign example as the starting point of the iterative attack.
Some methods consider the task from the perspective of optimization. In DeepFool~\cite{moosavi2016deepfool}, an optimized method is employed to generate the smallest perturbation while meeting the target of a successful attack. C\&W~\cite{carlini2017towards} transforms the task into a constrained optimization problem and compares the effects of multiple objective functions.

\textbf{Score-based Black-box Attack.}
This category of attack methods assumes they can obtain the classification probabilities of a given input image from the target model.
ZOO~\cite{chen2017zoo} proposes a zeroth-order optimization-based method, which directly estimates the gradients of the target model for generating adversarial examples.
However, it suffers from a low attack success rate and poor query efficiency since it is non-trivial to estimate the gradient with limited information. To address these problems, P-RGR~\cite{cheng2019improving} utilizes the transfer-based prior and query information to improve query efficiency.
$\mathcal{N}$ ATTACK~\cite{li2019nattack} further uses a probability density distribution over a small region centered around the input to generate adversarial examples, which defeats both vanilla DNNs and those generated by various defense techniques developed.

\textbf{Decision-based Black-box Attack.}
Under the decision-based black-box attack setting, only the predicted class of a given input image from the target model is available, which seems more difficult than the score-based black-box attack.
Boundary Attack~\cite{brendel2017decision} is firstly proposed to this problem, which gradually minimizes the adversarial perturbation by the approach of random walking while maintaining the aggressiveness of the adversarial example. Cheng \etal~\cite{cheng2018query} transforms the problem into a continuous real-valued optimization problem that can be solved by any zeroth-order optimization algorithm. Different from previous methods, Dong \etal~\cite{dong2019efficient} proposes an evolutionary attack algorithm, which reduces the dimension of the search space and improves the query efficiency.

\textbf{Transfer-based Black-box Attack.}
The transfer-based black-box attack cannot obtain any information of the target model, which is the most challenging setting. Existing methods~\cite{dong2018boosting, xie2019improving, dong2019evading, zou2020, lin2019nesterov} mainly endeavor to improve the attack success rates by the transferability of the adversarial examples generated from the substitute model. We will review these methods in \secref{sec:gradient} in details.
% MI-FGSM~\cite{dong2018boosting} is based on BIM and uses the momentum term to make the update direction in the iterative process more stable.
% DIM~\cite{xie2019improving} performs random resize and padding operations on the input image, which improves the transferability of adversarial examples. Compared with MI-FGSM, although it makes progress in black-box attacks, DIM slightly declines in the success rate of white-box attacks.
% TIM~\cite{dong2019evading} performs Gaussian convolution on the gradient information obtained at each step in the iterative process, which further improves the transferability of adversarial examples.
% By Integrated with DIM and TIM, Zou \etal~\cite{zou2020} explores the relationship between existing attack methods and proposes a multi-scale version attack method. Besides, NI-SI~\cite{lin2019nesterov} considers the problem from the perspective of optimization. This work uses Nesterov accelerated gradient to optimize the direction of updating of adversarial examples in the iteration, and proposes the scale-invariant attack method, which uses images of different pixel scales to generate adversarial examples.

Although these methods have been proposed to gradually improve the transferability of adversarial examples, the results in black-box attacks still leave a lot of room for improvement. Our work solves this problem from the perspective of meta-learning, which reduces the gap between white-box and black-box attacks, thus achieving better results.

\subsection{Adversarial Defense}
The task of adversarial defense is to improve the robustness of the model so that the model can correctly classify the perturbed adversarial examples. Defense methods can be divided into five categories: adversarial training, input transformation, randomization, model ensemble, and certified defenses.
Adversarial training~\cite{tramer2017ensemble, madry2017towards, song2019robust} conducts the model training based on the generated adversarial examples.
Input transformation utilizes JPEG compression~\cite{dziugaite2016study}, denoising~\cite{liao2018defense}, and extra GAN model~\cite{samangouei2018defense} before feeding the images to the model. Randomization refers to adding random noise into the input examples~\cite{xie2017mitigating} or models~\cite{dhillon2018stochastic} to make the model more robust to adversarial examples.
The model ensemble means the ensemble of multiple models in the output layer. Compared with a single model, it can reduce the impact of distributions in the adversarial examples~\cite{kurakin2018adversarial, liu2018towards}.
In addition, some work~\cite{raghunathan2018certified, wong2018scaling} proves that under a specific target model, a certified defense model can ensure robustness to adversarial examples.

\subsection{Meta-learning}
Meta-learning is a concept of learning how to learn. It mainly focuses on how to use a large amount of data to learn a learning mode. When encountering a new task later, only a small amount of additional data is needed to quickly complete the new task by fine-tuning the model~\cite{finn2017model, nichol2018first}.
As a newly proposed method in recent years, meta-learning has been widely used in various tasks. 
Meta-Attack~\cite{du2019query} uses the meta-learning framework to train the gradient estimator. In the case of the score-based black-box attack, only a small number of queries are needed to fine-tune the gradient estimator to complete the adversarial attack.

Different from Meta-Attack, our proposed MGAA is actually not a meta-learning framework but only inspired by its philosophy. Specifically, we do not need to explicitly train an extra model but directly generate the adversarial examples in the process of iteration.

%-------------------------------------------------------------------------
\section{Methodology}
Since our proposed architecture is plug-and-play and can be integrated with any existing gradient-based attack method, we first introduce some typical gradient-based attack methods in \secref{sec:gradient}. And then, we introduce our proposed Meta Gradient Adversarial Attack architecture in detail in \secref{sec:architecture}. Moreover, we give a detailed analysis of the effectiveness of our proposed architecture theoretically in \secref{sec:analysis}. Finally, a discussion on the differences between our proposed MGAA and some related works is given in \secref{sec:discussion}.

Let $x$ denote the benign example and $y$ denote the true label corresponding to the benign example. $f(x): \mathcal{X} \rightarrow \mathcal{Y}$ denotes the target classifier, $J$ denotes the loss function used for training the classifier, which usually means cross entropy loss. $x^{adv}$ denotes the adversarial example that needs to be optimized. The optimization objective can be described as: 
\begin{equation}
   \mathop{\arg\max}_{x^{adv}} J(f(x^{adv}), y), \quad \mathrm{s.t.} \|x^{adv} - x\|_{\infty} \leq \epsilon,
\end{equation}
where $\epsilon$ is the maximum perturbation per pixel allowed to add in the benign example. The goal of the adversarial example is to mislead the classifier into misclassification ($f(x^{adv} \neq y)$) with less perturbation added in compared to the benign example.

\subsection{Gradient-Based Attack Method} \label{sec:gradient}
We briefly introduce several gradient-based attack methods in this section, which focus on solving the problem of the white-box or transfer-based black-box attack.

\textbf{Fast Gradient Sign Method.} 
FGSM~\cite{goodfellow2014explaining} is a one-step method for white-box attack, which update the adversarial example in the direction of maximizing the loss function: 
\begin{equation}
   x^{adv} = x + \epsilon \cdot \mathrm{sign} (\nabla_x J(f(x), y)).
\end{equation}

\textbf{Basic Iterative Method.} 
Extended by FGSM, BIM~\cite{kurakin2016adversarial} proposes an iteration method to improve the success rate of white-box attack, which can be described as: 
\begin{equation}
   x^{adv}_{t+1} = x^{adv}_t + \alpha \cdot \mathrm{sign} (\nabla_x J(f(x^{adv}_t), y)),
\end{equation}
where $x^{adv}_0\! = \!x$, $\alpha\! =\! \epsilon / T$ and $T$ is the number of iterations.

\textbf{Momentum Iterative Fast Gradient Sign Method.}
A momentum term is further proposed in MIM~\cite{dong2018boosting} to accumulate the update direction of previous steps in the process of iteration:
\begin{gather}
   g_{t+1} = \mu \cdot g_t + \frac{\nabla_x J(f(x^{adv}_t), y)}{\| \nabla_x J(f(x^{adv}_t), y)\|_1}, \\
   x^{adv}_{t+1} = x^{adv}_t + \alpha \cdot \mathrm{sign}(g_{t+1}),
\end{gather}
where $g_t$ denotes the momentum term of gradient in the $t$-th iteration and $\mu$ is a decay factor.
Compared with BIM, MIM achieves significant progress in the black-box attack.

\textbf{Diverse Inputs Method.} 
DIM~\cite{xie2019improving} proposes to apply random resizing and padding transformations in the benign example to improve the transferability of adversarial examples. Although this method is simple and easy to implement, it also brings great improvements.

\textbf{Translation-Invariant Attack Method.} 
Different from previous methods, TIM~\cite{dong2019evading} proposes to convolve the gradient with a Gaussian kernel, which can be combined with MIM:
\begin{gather}
   g_{t+1} = \mu \cdot g_t + \frac{W \ast \nabla_x J(f(x^{adv}_t), y)}{\|W \ast \nabla_x J(f(x^{adv}_t), y)\|_1}, \\
   x^{adv}_{t+1} = x^{adv}_t + \alpha \cdot \mathrm{sign}(g_{t+1}),
\end{gather}
where $W$ is a Gaussian kernel and $\ast$ denotes the operator of convolution.

\textbf{SI-NI.}
Moreover, SI-NI~\cite{lin2019nesterov} proposes two approaches to further improve the transferability of adversarial examples, \ie, NI-FGSM and SIM. NI-FGSM integrates Nesterov Accelerated Gradient~\cite{Nesterov1983AMF} into the iterative gradient-based attack, \eg MIM, to conduct a robust adversarial attack:
\begin{gather}
   x^{nes}_t = x^{adv}_t + \alpha \cdot \mu \cdot g_t, \\
   g_{t+1} = \mu \cdot g_t + \frac{\nabla_x J(f(x^{nes}_t), y)}{\| \nabla_x J(f(x^{nes}_t), y)\|_1}, \\
   x^{adv}_{t+1} = x^{adv}_t + \alpha \cdot \mathrm{sign}(g_{t+1}).
\end{gather}
SIM optimizes the adversarial examples over the scale copies of the benign example:
\begin{equation}
   \begin{split}
      \mathop{\arg\max}_{x^{adv}} \frac{1}{m} \sum\nolimits_{i=0}^m J(f(S_i(x^{adv})), y), \\
      \mathrm{s.t.} \|x^{adv} - x\|_{\infty} \leq \epsilon,
   \end{split}
\end{equation}
where $S_i(x)=x/2^i$ denotes the scaled image and $m$ denotes the number of scale copies.

\subsection{Meta Gradient Adversarial Attack} \label{sec:architecture}
Most of the existing transfer-based black-box attack methods use gradient-based methods to generate adversarial examples against the white-box models, and then directly utilize the transferability of the adversarial examples to conduct the attack against the black-box model. However, due to the differences in the structures and the parameters between different models, relying only on the transferability of adversarial examples to conduct the black-box attack cannot achieve perfect results, which leaves a lot of room for improvement.

In this paper, we solve the problem of the black-box attack from a different perspective. Inspired by the philosophy of meta-learning, we propose an architecture called Meta Gradient Adversarial Attack (MGAA), which is plug-and-play and can be integrated with any existing gradient-based attack method. As shown in \figref{fig:architecture}, $T$ tasks are sampled from a model zoo iteratively. In each task, we simulate a white-box attack in the meta-train step and a black-box attack in the meta-test step, narrowing the gap of gradient directions between white-box and black-box settings to improve the transferability.

Specifically, suppose there are a total of $N$ models of $M_1, M_2, \cdots, M_N$ in the model zoo, we randomly select $n+1$ models to compose a task in each iteration, which consists of two steps, \ie, meta-train and meta-test. At the beginning of each iteration (take $i$-th iteration as an example), we take the adversarial example generated in the previous iteration $x_i$ as the input, denoted as $x_{i,0}$.

\textbf{Meta-train.} 
A total of $n$ models of $M_{k_1}$, $M_{k_2}$, $\cdots$, $M_{k_{n}}$ are utilized to simulate the white-box attack. We employ the same approach as~\cite{dong2018boosting} when doing the adversarial attack by the ensemble of multiple models. Specifically, to attack the ensemble of $n$ models, we fuse the logits as:
\begin{equation}
   l(x_{i,0}) = \sum\nolimits_{s=1}^{n} w_s l_{k_s}(x_{i,0}),
\end{equation}
where $l_{k_s}(x_{i,0})$ is the logits of the model $M_{k_s}$, $w_s$ is the ensemble weight of each model with $w_s \geq 0$ and $\sum\nolimits_{s=1}^{n} w_s=1$. Then the cross entropy loss is used to calculate the loss of misclassification:
\begin{equation}
   \label{equ:train_loss}
   \mathcal{L}_{M_{k_1}, \cdots, M_{k_{n}}}(x_{i,0})= -\mathbbm{1}_y  \cdot \text{log}(\text{softmax}(l(x_{i,0}))),
\end{equation}
where $\mathbbm{1}_y$ is the one-hot encoding of $y$.
Same as the common method of gradient-based attack, the adversarial example is updated along the direction of maximizing the loss function:
\begin{equation}
   x_{i,j+1}=x_{i,j}+\alpha \cdot \text{sign}(\nabla_{x_{i,j}}\mathcal{L}_{M_{k_1}, \cdots, M_{k_{n}}}(x_{i,j})),
\end{equation}
where $\alpha$ is the step size in meta-train step. The meta-train step can be iterated for $K$ times, and the subscript $j$ denotes the number of iteration in meta-train step. It is worth noting that our proposed architecture can be integrated with any gradient-based attack methods, and we only take the formula of update in BIM~\cite{kurakin2016adversarial} as an example here for convenience. 

\textbf{Meta-test.} 
After using the ensemble of multiple models to simulate the white-box attack, we use the last sampled model $M_{k_{n+1}}$ to simulate the black-box attack on the basis of generated adversarial example $x_{i,K}$ in the meta-train step, where $K$ is the number of iteration in meta-train step.

Specifically, we first calculate the cross entropy loss by model $M_{k_{n+1}}$:
\begin{equation}
   \label{equ:test_loss}
   \mathcal{L}_{M_{k_{n+1}}}(x_{i,K})= -\mathbbm{1}_y \cdot \text{log}(\text{softmax}(l_{k_{n+1}}(x_{i,K}))),
\end{equation}
where $l_{k_{n+1}}(x_{i,K})$ is the logits of model $M_{k_{n+1}}$.
We then update the adversarial example based on $x_{i,K}$ along the direction of maximizing the loss function in meta-test step:
\begin{equation}
   x_{i,mt}=x_{i,K}+\beta \cdot \text{sign}(\nabla_{x_{i,K}}\mathcal{L}_{M_{k_{n+1}}}(x_{i,K})),
\end{equation}
where $\beta$ is the step size in meta-test step.

To improve the transferability of the adversarial example, as shown in \figref{fig:architecture}, we add the perturbation obtained in the meta-test step to the adversarial example generated in the previous iteration $x_i$ to update the adversarial example:
\begin{equation}
   x_{i+1} = x_{i} + (x_{i,mt} - x_{i,K}).
\end{equation}

The iteration of meta-train and meta-test can be conducted by a total of $T$ times to generate the final adversarial example, where each iteration $i$ randomly picks different models to establish various tasks and takes the output of the previous iteration $i-1$ as input. The procedure of MGAA is summarized in \algref{algorithm}.

It deserves to mention that the (n+1)-th model is not the black-box model in the real testing scenario. Actually, it is just a simulated black-box model to adaptively narrow the gaps of gradient directions between the meta-train and meta-test steps by iteratively simulating the white-box and black-box attacks.

\begin{algorithm}[t]
   \caption{Meta Gradient Adversarial Attack}
   \label{algorithm}
   \begin{algorithmic}[1]
      \REQUIRE \quad \\the input example $x_1$\\the classifier models $M_1, M_2, \cdots, M_N$%\\ the number of sample tasks $T$\\ the iteration number in meta-train step $K$\\ the step size in meta-train step $\alpha$\\ the step size in meta-test step $\beta$
      \ENSURE the adversarial example $x_{T+1}$ 
      \FOR {i $\in \{1, \cdots T\}$ }
         \STATE Randomly sample $n+1$ models $M_{k_1}$, $M_{k_2}$, $\cdots$, $M_{k_{n+1}}$ from $M_1$, $M_2$, $\cdots$, $M_N$ as a task  \label{alg:task}
         \STATE $x_{i,0} = x_i$  
         \FOR {j $\in \{0, 1, \cdots K-1\}$ }
            \STATE Calculate the cross entropy loss of input $x_{i,j}$ under the ensemble of $n$ models $M_{k_1}, M_{k_2}, \cdots, M_{k_{n}}$: $\mathcal{L}_{M_{k_1}, \cdots, M_{k_{n}}}(x_{i,j}) $ as \equref{equ:train_loss}
            \STATE $x_{i,j+1}=x_{i,j}+\alpha \cdot \text{sign}(\nabla_{x_{i,j}}\mathcal{L}_{M_{k_1}, \cdots, M_{k_{n}}}(x_{i,j}))$
         \ENDFOR
         \STATE Calculate the cross entropy loss of input $x_{i,K}$ under the model $M_{k_{n+1}}$: $\mathcal{L}_{M_{k_{n+1}}}(x_{i,K})$ as \equref{equ:test_loss}
         \STATE $x_{i,mt}=x_{i,K}+\beta \cdot \text{sign}(\nabla_{x_{i,K}}\mathcal{L}_{M_{k_{n+1}}}(x_{i,K}))$
         \STATE $x_{i+1}=x_i+(x_{i,mt}-x_{i,K})$
      \ENDFOR
   \end{algorithmic}
   % \vspace{-1mm}
\end{algorithm}

% 这段话到底放在这里还是放在discussion里面
% In the iterative process, different combinations of models from the model zoo are sampled to compose different tasks, which help the adversarial example access the various model distributions when they are generated. At the same time, unlike the existing transfer-based methods, which only conduct the white-box attack all the time (similar to the meta-train step in our proposed architecture), Meta Gradient Adversarial Attack proposes to use another model to further simulate the black-box attack in the meta-test step. And only the perturbation obtained in the meta-test step is actually added to the adversarial example generated in the last iteration of the task. This can well narrow the gap between gradient directions in the white-box attack and the black-box attack. Thus the whole architecture can be seen as iteratively conducting the simulated black-box attack, which can improve the transferability and the attack success rate of generated adversarial examples in the setting of the black-box attack test.

\subsection{Analysis} \label{sec:analysis}
In this section, we give a detailed analysis on the reasons why our proposed architecture is effective theoretically. To be specific, we consider the objective function in the meta-test step:
\begin{equation}
   \label{equ:raw}
   \mathop{\arg\max}_{x} J(f_{n+1}(x+\alpha \nabla_{x} J(\sum\nolimits_{i=1}^{n}f_i(x)/n, y)), y),
\end{equation}
where $J$ denotes the cross entropy loss, $f_i$ denotes the $i$-th sampled model in the task, $n$ denotes the number of models used in the meta-train step, $f_{n+1}$ denotes the sampled model used in the meta-test step, $x$ and $y$ denote the adversarial example and the true class label of the benign example respectively.
Similar to~\cite{li2017learning}, with the first order Tayler expansion of \equref{equ:raw} at point $x$, the objective function can be rewritten as:
\begin{align}
   \label{equ:full}
   \mathop{\arg\max}_{x} &J(f_{n+1}(x), y) +  \\
                         &\alpha \nabla_{x} J(\sum\nolimits_{i=1}^{n}f_i(x)/n, y) \nabla_{x} J(f_{n+1}(x), y). \notag
\end{align}

The first term in \equref{equ:full} can be regarded as the optimization of the meta-test step, and the second term can be regarded as the calculation of cosine similarity of two gradients, that is, constraining the gradient directions in meta-train and meta-test steps as similar as possible, which is consistent with our motivation of narrowing the gaps of the gradient directions between them to improve the transferability of generated adversarial examples. We also conduct experiments to verify that the directions of the generated adversarial perturbations by our proposed MGAA are more similar to the gradients of the black-box model than existing methods, which are provided in \secref{sec:cosine}.

The existing gradient-based methods only focus on the update of the gradient during the white-box attack, and cannot obtain any information about the black-box attack. In contrast, our proposed architecture iteratively simulates the white-box attack and black-box attack, and constrains the gradient directions of the two to be similar in each iteration. In the iterative process, different combinations of models from the model zoo are sampled to compose different tasks, which help the adversarial examples access various model distributions when they are generated. The adversarial examples obtained after multiple iterations may not be task-specific optimal, \ie, biased to any existing white-box models, but it generally contains better transferability for any unseen black-box model. This explains why our proposed architecture can improve the success rates of both white-box and black-box attacks.

\subsection{Discussion} \label{sec:discussion}
\textbf{Differences from Meta-Attack~\cite{du2019query}.} In spite of the similarity in name, our proposed Meta Gradient Adversarial Attack (MGAA) is quite different from Meta-Attack~\cite{du2019query}. Firstly, the tasks solved by these two methods are different. Meta-Attack solves the task of the score-based black-box attack, while our proposed MGAA addresses the task of the transfer-based black-box attack, which is more challenging. More importantly, Meta-Attack trains an extra gradient estimator as vanilla meta-learning methods do to estimate the gradient information of different input examples. Since the gradient estimator is a generative model, it can be quite difficult to learn. In contrast, no extra model is needed to train in our proposed MGAA architecture, which only borrows the idea of meta-learning and directly updates the adversarial examples through gradients in the iterative process. By easily integrating with the gradient-based methods, which are dominating the current black-box attack field, in a plug-and-play mode, our MGAA architecture has greater potential to generate more aggressive adversarial examples.

\textbf{Differences from existing ensemble strategies.} There have been several ensemble-based adversarial attack methods. We give a brief discussion on the differences between our MGAA and them. Liu \etal~\cite{liu2016delving} proposes an optimization-based method to generate adversarial examples by the ensemble of models, which is totally different from our framework integrated with the gradient-based attack method. MIM~\cite{dong2018boosting} proposes an efficient ensemble strategy for gradient-based attack methods. By analyzing three kinds of ensemble strategies of the models, MIM finds that ensemble the results of models by the logits layer is the most effective way to generate adversarial examples. Different from the simple way to summarize the logits layer of all the models, we propose a more advanced architecture, which gradually narrows the gap of update directions between white-box attack and black-box attack settings to further improve the transferability of adversarial examples.

%-------------------------------------------------------------------------
\section{Experiment}
In this section, we conduct extensive experiments to verify the effectiveness of the proposed Meta Gradient Adversarial Attack architecture. We first introduce the settings in the experiments, including the datasets, models, and experiment details in \secref{sec:setting}. Then some experiments are performed to investigate the impact of different hyperparameters in the Meta Gradient Adversarial Attack architecture in \secref{sec:hyperparameter}. Moreover, our proposed architecture is compared with the state-of-the-art methods, demonstrating the superiority of our Meta Gradient Adversarial Attack method under both the targeted and untargeted setting in \secref{sec:sota}.
The ablation study of our MGAA is provided in \secref{sec:ablation}, which demonstrates the effects of the meta-train and the meta-test step, respectively. Moreover, the experiments of attack under various perturbation budgets and the mininum adversarial noises used to attack are provided in \secref{sec:budget} and \secref{sec:min_noise}, respectively.

\subsection{Settings}
\label{sec:setting}
\textbf{Datasets.}
We use ImageNet~\cite{russakovsky2015imagenet} and CIFAR10~\cite{krizhevsky2009learning} datasets to conduct the experiments. For ImageNet, the ImageNet-compatible dataset\footnote{\url{https://github.com/tensorflow/cleverhans/tree/master/examples/nips17_adversarial_competition/dataset}}~\cite{russakovsky2015imagenet} in the NIPS 2017 adversarial competition is used, which contains 1000 images with a resolution of $299 \times 299 \times 3$. A baseline model of Inceptionv3~\cite{szegedy2016rethinking} can achieve a classification accuracy of 100\% on these images. For CIFAR10, the test set with 10000 images is evaluated in our experiments.

\textbf{Models.}
Our architecture utilizes a total of 10 white-box models to generate adversarial examples. In each iteration, multiple models are randomly selected to compose a meta-task. Under a black-box attack scenario, we evaluate 6 and 7 models on ImageNet and CIFAR10, respectively. All the models used in white-box and black-box settings are shown in \tabref{tab:model}. The details of these models are provided in \secref{sec:models}.
\begin{table}[t]
   \begin{center}
      \caption{Models on CIFAR10 and ImageNet. The first 10 models are treated as white-box models and the rest models are black-box models in our experimental settings.}
      \label{tab:model}
      \vspace{-4mm}
      \resizebox{\columnwidth}{!}{
         \begin{tabular}{c|c|c|c}
            \hline 
            ~ & No. & ImageNet & CIFAR10 \\
            \hline
            \multirow{10}*{\begin{tabular}{c}white-\\box\\models\end{tabular}} & 1 & Inceptionv3~\cite{szegedy2016rethinking} & ResNet-18~\cite{he2016deep}\\ \cline{2-4}
            ~ & 2 & Inceptionv4~\cite{szegedy2016inception} & ResNetv2-18~\cite{he2016identity}\\ \cline{2-4}
            ~ & 3 & InceptionResNetv2~\cite{szegedy2016inception} & GoogLeNet~\cite{szegedy2015going}\\ \cline{2-4}
            ~ & 4 & ResNetv2-152~\cite{he2016identity} & ResNeXt-29~\cite{xie2017aggregated}\\ \cline{2-4}
            ~ & 5 & Ens3\_Inceptionv3~\cite{tramer2017ensemble} & SENet-18~\cite{hu2018squeeze}\\ \cline{2-4}
            ~ & 6 & Ens4\_Inceptionv3~\cite{tramer2017ensemble} & RegNetX-200mf~\cite{radosavovic2020designing} \\ \cline{2-4}
            ~ & 7 & Ens\_InceptionResNetv2~\cite{tramer2017ensemble} & DLA~\cite{yu2018deep}\\ \cline{2-4}
            ~ & 8 & ResNetv2-101~\cite{he2016identity} & Shake-ResNet-26\_2x64d~\cite{gastaldi2017shake}\\ \cline{2-4}
            ~ & 9 & MobileNetv2\_1.0~\cite{sandler2018mobilenetv2} & Adv\_ResNet-18\\ \cline{2-4}
            ~ & 10 & PNasNet~\cite{liu2018progressive} & Adv\_DenseNet-121\\ 
            \hline
            \multirow{7}*{\begin{tabular}{c}black-\\box\\models\end{tabular}} & 11 & Adv\_Inceptionv3~\cite{tramer2017ensemble} & PyramidNet-164~\cite{han2017deep}\\ \cline{2-4}
            ~ & 12 & NasNet\_mobile~\cite{zoph2018learning} & CbamResNeXt~\cite{woo2018cbam} \\ \cline{2-4}
            ~ & 13 & MobileNetv2\_1.4~\cite{sandler2018mobilenetv2} & Adv\_GoogLeNet \\ \cline{2-4}
            ~ & 14 & R\&P~\cite{xie2017mitigating} & Adv\_ResNet-18\_ll \\ \cline{2-4}
            ~ & 15 & NIPS-r3 & $k$-WTA~\cite{xiao2019enhancing} \\ \cline{2-4}
            ~ & 16 & CERTIFY~\cite{cohen2019certified} & GBZ~\cite{li2019generative} \\ \cline{2-4}
            ~ & 17 & ~ & ADP~\cite{pang2019improving} \\
            \hline
         \end{tabular}
      }
   \end{center}
   \vspace{-5mm}
\end{table}

\begin{table*}[t]
   \begin{center}
      \caption{The attack success rates of the adversarial examples generated under \textbf{different number of iterations $K$} in meta-train step against the white-box and black-box models on ImageNet. The number of sampled tasks $T$ is 40. The number of ensembled models $n$ in the meta-train step is 5. The index of models in the table is the same as \tabref{tab:model}.} %The index of models in the table is the same as \tabref{tab:model}.}
      \vspace{-4mm}
      \label{tab:iter}
      \resizebox{0.95\textwidth}{!}{
         \begin{tabular}{c|c|c|c|c|c|c|c|c|c|c|c|c|c|c|c|c}
            \hline 
            \multirow{2}*{$K$} & \multicolumn{10}{c|}{white-box models} & \multicolumn{6}{|c}{black-box models} \\ \cline{2-17}
            & 1 & 2 & 3 & 4 & 5 & 6 & 7 & 8 & 9 & 10 & 11 & 12 & 13 & 14 & 15 & 16\\ \hline
            1 & 99.1 & 99.2 & 97.7 & 98.1 & 98.7 & 98.8 & 97.2 & 98.1 & 98.6 & 98.5 &  96.3 & 97.7 & 97.5 & 95.9 & 96.3 & 68.2 \\
            2 & 99.5 & 99.5 & 98.9 & 98.5 & 99.2 & 99.0 & 98.2 & 98.6 & 99.0 & 98.9 &  97.4 & 98.2 & 98.3 & 97.7 & 97.6 & 70.2 \\
            5 & 99.9 & \textbf{100} & 99.7 & 99.5 & 99.8 & 99.7 & 98.9 & 99.5 & 99.5 & 99.7 &  98.6 & 99.3 & 99.1 & 98.7 & 98.6 & 71.3 \\
            8 & \textbf{100} & \textbf{100} & \textbf{99.9} & \textbf{99.8} & \textbf{99.9} & \textbf{99.8} & \textbf{99.4} & \textbf{99.8} & \textbf{99.6} & \textbf{100} & \textbf{99.1} & \textbf{99.4} & \textbf{99.5} & \textbf{99.4} & \textbf{99.3} & \textbf{71.6} \\
            \hline
         \end{tabular}
      }
   \end{center}
   \vspace{-5mm}
\end{table*}

\begin{table*}[!t]
   \begin{center}
      \caption{The success rates under \textbf{targeted attack} setting on \textbf{ImageNet}. The number of ensembled models $n$ in meta-train step is 5.  The number of iterations $K$ in meta-train step is 5. }
      \vspace{-4mm}
      \label{tab:target}
      \resizebox{\textwidth}{!}{
         \begin{tabular}{c|c|c|c|c|c|c|c|c|c|c|c|c|c|c|c|c|c}
            \hline
            \multirow{2}*{$n$} & \multicolumn{11}{c|}{white-box models} & \multicolumn{6}{|c}{black-box models} \\ \cline{2-18}
            & 1 & 2 & 3 & 4 & 5 & 6 & 7 & 8 & 9 & 10 & avg. & 11 & 12 & 13 & 14 & 15 & avg. \\ 
            \hline
            MIM~\cite{dong2018boosting} & 96.7 &80.8 &68.3 &59.7 &96.2 &97.0 &78.0 &65.5 &77.1 &38.5 &75.78 &0.0 &6.9 &6.8 &0.3 &0.2 &2.84 \\
            MGAA w/ MIM & \textbf{99.7} &\textbf{99.2} &\textbf{96.9} &\textbf{93.5} &\textbf{99.2} &\textbf{99.7} &\textbf{96.9} &\textbf{94.9} &\textbf{98.6} &\textbf{84.7} &\textbf{96.33} &0.0 &\textbf{19.9} &\textbf{19.1} &\textbf{1.8} &\textbf{2.0} &\textbf{8.56} \\
            \hline
            DIM~\cite{xie2019improving} & 84.7 &74.8 &70.1 &59.7 &74.2 &76.2 &47.9 &60.8 &73.4 &52.5 &67.43 &0.8 &32.4 &30.4 &16.7 &16.8 &19.42 \\
            MGAA w/ DIM & \textbf{99.4} &\textbf{98.3} &\textbf{96.3} &\textbf{92.5} &\textbf{96.4} &\textbf{97.5} &\textbf{87.9} &\textbf{94.4} &\textbf{97.3} &\textbf{91.0} &\textbf{95.10} &\textbf{3.5} &\textbf{65.5} &\textbf{62.7} &\textbf{54.2} &\textbf{53.6} &\textbf{47.90} \\
            \hline
            TI-DIM~\cite{dong2019evading} & 54.8 &46.2 &39.0 &47.2 &39.5 &40.1 &32.1 &20.3 &49.6 &40.4 &42.85 &\textbf{39.6} &23.7 &14.5 &34.2 &26.6 &23.86 \\
            MGAA w/ TI-DIM & \textbf{96.0} &\textbf{90.3} &\textbf{82.8} &\textbf{88.6} &\textbf{82.2} &\textbf{83.0} &\textbf{71.7} &\textbf{91.5} &\textbf{88.7} &\textbf{80.4} &\textbf{85.52} &37.3 &\textbf{40.7} &\textbf{24.5} &\textbf{44.4} &\textbf{44.0} &\textbf{38.18} \\
            \hline
         \end{tabular}
      }
   \end{center}
   \vspace{-5mm}
\end{table*}

\textbf{Experiment Details.}
The range of pixel value in each image is 0-255, and our maximum perturbation $\epsilon$ is set to 16 on ImageNet and 8 on CIFAR10. We compare our proposed architecture with MIM~\cite{dong2018boosting}, DIM~\cite{xie2019improving}, TIM~\cite{dong2019evading} and SI-NI~\cite{lin2019nesterov} methods. For MIM, we adopt the decay factor $\mu=1$. For DIM, the transformation probability is set to $0.7$. For TIM, the size of the Gaussian kernel is set to $7\times 7$. For SI-NI, the number of scale copies is set to $m=5$. For all iterative methods, including ours, the number of iteration $T$ is set to 40. Unless mentioned, all experiments in this section are based on the integration of TI-DIM~\cite{dong2019evading} method with our proposed architecture. The step sizes in meta-train step $\alpha$ and meta-test step $\beta$ are $1$ and $\epsilon/T$, respectively. All the experiments of baseline methods~\cite{xie2019improving, dong2019evading, lin2019nesterov} in our paper all adopt the ensemble strategy in MIM~\cite{dong2018boosting} since it is the most effective strategy available.

\subsection{Impact of Hyperparameters}
\label{sec:hyperparameter}
In our proposed architecture of Meta Gradient Adversarial Attack (MGAA), three hyperparameters may affect attack success rates: the number of models $n$ selected for the ensemble-based attacks during the meta-train step, the number of iterations $K$ during the meta-train step, and the number of tasks $T$ sampled during the entire generation. We analyze in detail how does each hyperparameter affect the final performance of our proposed architecture. Due to the space limitations, we only elaborate the analysis of $K$ here. The results and detailed analysis of $n$ and $T$ are provided in \secref{sec:hyperparameters}.

\textbf{The number of iterations $K$ in the meta-train step.}
In the meta-train step, the number of iteration steps $K$ used to iteratively updating the adversarial example plays an important role in improving the success rates by the ensemble of multiple models. We compare the attack success rates of generated adversarial examples against white-box and black-box models with different iteration steps $K$ in \tabref{tab:iter}. With more iteration steps, the generated adversarial examples are more aggressive. When these aggressive adversarial examples are used as the basis for the meta-test step, the transferability of perturbation obtained by the meta-test step can also be stronger. But at the same time, the more iteration steps mean the more time it takes to generate adversarial examples. We recommend the value of $K$ to be 5 in the following experiments.

\textbf{The number of sampled tasks $T$.}
The more sampled tasks are taken, the higher attack success rate can be achieved, especially for black-box settings. However, on the other hand, increasing the number of sampled tasks also increases the time it takes to generate adversarial examples. Considering a trade-off of both efficiency and effectiveness, the value of $T$ is recommended to be 40. The detailed results are provided in \secref{sec:hyperparameters}.

\textbf{The number of ensembled models $n$ in meta-train step.}
When the number of ensembled models increases, the success rates against the white-box and black-box attacks become higher and higher. But when $n$ is greater than 5, the increase in attack success rates is not obvious. Considering that the more ensembled models in each iteration, the higher the computational complexity is needed, the number of ensembled models being 5 is a suitable choice. The detailed results are provided in \secref{sec:hyperparameters}.

\subsection{Compared with the State-of-the-art Methods}
\label{sec:sota}
\subsubsection{The Targeted Attack}
The experiments of adversarial examples generated on ImageNet under the targeted attack setting are shown in \tabref{tab:target}. The target label of each image is provided by the dataset. Compared to the baseline methods, the attack success rate significantly increases when integrating baseline methods with our proposed MGAA. For the DIM method~\cite{xie2019improving}, the integration with MGAA brings an average increase of 27.67\% and 28.52\% in white-box and black-box settings, respectively, reaching 95.10\% and 47.90\% respectively.

\subsubsection{The Untargeted Attack}
\begin{table*}[!hbtp]
   \begin{center}
      \caption{The success rates under \textbf{untargeted attack} setting on \textbf{ImageNet}. The number of ensembled models $n$ in meta-train step is 5.  The number of iterations $K$ in meta-train step is 8.} %The index of models in the table is the same as \tabref{tab:model}.}
      \vspace{-3mm}
      \label{tab:sota_imagenet}
      \resizebox{\textwidth}{!}{
         \begin{tabular}{c|c|c|c|c|c|c|c|c|c|c|c|c|c|c|c|c|c}
            \hline 
            \multirow{2}*{Method} & \multicolumn{10}{c|}{white-box models} & \multicolumn{6}{|c|}{black-box models} & Time \\ \cline{2-17}
            & 1 & 2 & 3 & 4 & 5 & 6 & 7 & 8 & 9 & 10 & 11 & 12 & 13 & 14 & 15 & 16 & ($s$/img)\\ \hline
            SI-NI~\cite{lin2019nesterov}  & 99.7 &97.5 &97.4 &96.3 &98.8 &98.6 &90.8 &95.6 &99.7 &98.2 &48.2 &90.8 &92.9 &50.6 &58.5 &38.9 &68.29 \\ \hline
            MIM~\cite{dong2018boosting}   & 99.6 &99.7 &99.4 &98.7 &99.8 &99.8 &99.5 &99.0 &99.1 &98.2 &44.4 &92.6 &94.1 &65.4 &72.2 &34.4 &17.51 \\
            MGAA w/ MIM                   & \textbf{100} &\textbf{100} &\textbf{100} &\textbf{99.9} &\textbf{100} &\textbf{100} &\textbf{100} &\textbf{99.9} &\textbf{99.9} &\textbf{99.9} &\textbf{52.0} &\textbf{96.0} &\textbf{96.9} &\textbf{67.1} &\textbf{74.9} &\textbf{37.0} &71.24 \\ \hline
            DIM~\cite{xie2019improving}   & 99.4 &99.8 &99.5 &98.6 &99.4 &99.5 &98.5 &98.6 &98.9 &98.8 &79.4 &98.0 &98.3 &95.0 &95.3 &44.8 &22.22 \\
            MGAA w/ DIM                   & \textbf{100} &\textbf{100} &\textbf{100} &\textbf{99.9} &\textbf{100} &\textbf{100} &\textbf{99.9} &\textbf{99.9} &\textbf{100} &\textbf{99.9} &\textbf{88.0} &\textbf{99.9} &\textbf{99.8} &\textbf{98.9} &\textbf{98.9} &\textbf{49.3} &69.26 \\ \hline
            TI-DIM~\cite{dong2019evading} & 98.9 &99.1 &98.2 &98.3 &98.9 &98.6 &97.3 &98.0 &98.1 &98.3 &96.3 &97.5 &97.5 &96.7 &96.8 &67.8 &19.13 \\ 
            MGAA w/ TI-DIM                & \textbf{100} &\textbf{100} &\textbf{99.9} &\textbf{99.8} &\textbf{99.9} &\textbf{99.8} &\textbf{99.4} &\textbf{99.8} &\textbf{99.6} &\textbf{100} &\textbf{99.1} &\textbf{99.4} &\textbf{99.5} &\textbf{99.4} &\textbf{99.0} &\textbf{71.6} &67.28 \\ \hline
         \end{tabular}
      }
   \end{center}
   \vspace{-5mm}
\end{table*}

\begin{table*}[!hbtp]
   \begin{center}
      \caption{The success rates under \textbf{untargeted attack} setting on \textbf{CIFAR10}. The number of ensembled models $n$ in meta-train step is 5.  The number of iterations $K$ in meta-train step is 5.} %The index of models in the table is the same as \tabref{tab:model}.}
      \vspace{-3mm}
      \label{tab:sota_cifar10}
      \resizebox{\textwidth}{!}{
         \begin{tabular}{c|c|c|c|c|c|c|c|c|c|c|c|c|c|c|c|c|c}
            \hline 
            \multirow{2}*{Method} & \multicolumn{10}{c|}{white-box models} & \multicolumn{7}{|c}{black-box models}  \\ \cline{2-18}
            & 1 & 2 & 3 & 4 & 5 & 6 & 7 & 8 & 9 & 10 & 11 & 12 & 13 & 14 & 15 & 16 & 17 \\ \hline
            MIM~\cite{dong2018boosting}   & 99.84 &99.96 &99.99 &99.98 &99.94 &99.76 &\textbf{99.89} &99.72 &80.39 &96.68 &98.49 &99.56 &90.55 &39.20 &96.34 &82.92 &96.66 \\
            MGAA w/ MIM                   & \textbf{99.99} &\textbf{100} &\textbf{100} &\textbf{99.99} &\textbf{99.99} &\textbf{99.91} &99.87 &\textbf{99.87} &\textbf{85.86} &\textbf{99.19} &\textbf{99.35} &\textbf{99.97} &\textbf{97.49} &\textbf{55.94} &\textbf{97.09} &\textbf{87.20} &\textbf{97.69} \\ \hline
            DIM~\cite{xie2019improving}   & 99.86 &99.98 &99.98 &99.99 &99.98 &99.75 &99.90 &99.81 &80.71 &96.45 &99.03 &99.68 &93.34 &45.62 &96.38 &85.65 &96.95 \\
            MGAA w/ DIM                   & \textbf{99.98} &\textbf{100} &\textbf{100} &\textbf{100} &\textbf{99.99} &\textbf{99.87} &\textbf{99.99} &\textbf{99.93} &\textbf{85.07} &\textbf{98.88} &\textbf{99.48} &\textbf{99.89} &\textbf{98.71} &\textbf{68.22} &\textbf{97.26} &\textbf{90.07} &\textbf{98.84} \\ \hline
            TI-DIM~\cite{dong2019evading} & 99.86 &99.96 &99.98 &99.99 &99.97 &99.76 &99.94 &99.85 &81.17 &96.66 &99.27 &99.74 &93.61 &46.69 &96.50 &85.76 &96.92 \\ 
            MGAA w/ TI-DIM                & \textbf{99.95} &\textbf{100} &\textbf{100} &\textbf{100} &\textbf{100} &\textbf{99.93} &\textbf{99.98} &\textbf{99.90} &\textbf{85.63} &\textbf{99.19} &\textbf{99.51} &\textbf{99.99} &\textbf{98.76} &\textbf{68.77} &\textbf{97.35} &\textbf{90.04} &\textbf{98.62} \\ \hline
         \end{tabular}
      }
   \end{center}
   \vspace{-10mm}
\end{table*}

We compare our proposed MGAA with some typical gradient-based attack methods on ImageNet and CIFAR10 in \tabref{tab:sota_imagenet} and \tabref{tab:sota_cifar10} respectively under untargeted attack setting. Both results show the superiority of our proposed MGAA. It can be seen from \tabref{tab:sota_imagenet} that although the SI-NI method~\cite{lin2019nesterov} is effective in generating adversarial examples from a single model for the transfer-based black-box attack, the result is not satisfactory enough when using the ensemble of multiple models to conduct black-box attacks. Also, this method is time-consuming and memory-consuming due to the need for multiple scale copies of the model, which is unsuitable for the ensemble of multiple models. Thus we do not integrate it with our MGAA. Integrating other existing methods (\eg MIM, DIM, and TI-DIM) with our proposed MGAA architecture, the attack success rates against the white-box and black-box models have been consistently improved. %The experimental results show that our proposed architecture is quite effective and flexible, which can be integrated with gradient-based attack methods to further improve the attack success rates under both white-box and black-box settings. 

\tabref{tab:sota_imagenet} also gives the time cost of our MGAA on a GTX 1080Ti GPU.
Although MGAA performs relatively slower when integrated with the existing gradient-based attack method, our MGAA achieves higher attack success rates when the $T$ is 40. We also conduct the experiments with $T$ being 10 in \secref{sec:more}, and similar conclusions can be obtained. Further, when comparing our results with $T$ being 10 and the results of TI-DIM with $T$ being 40, we can see that the time consumed is nearly equal, but our method achieves higher attack success rates on both white-box and black-box settings.

% \vspace{-5mm}
% \subsubsection{Attack under Various Perturbation Budgets}
% We conduct experiments of the attack under various perturbation budgets. The curve of attack success rate vs. perturbation budgets on ImageNet is shown in \figref{fig:curve}. We can clearly see that TI-DIM with our MGAA achieves higher attack success rates in both white-box and black-box attacks, which further verify that our method can achieve consistently better performance under various perturbation budgets. More detailed results of each model are provided in the supplementary material.

% \begin{figure}
%    \centering
%    \includegraphics[width=\columnwidth]{Img/curve_avg.pdf}
%    \vspace{-2mm}
%    \caption{The attack success rates vs. perturbation budget curve on ImageNet. The dotted lines are the results of TI-DIM, and the solid lines are the results of our method.}
%    \label{fig:curve}
%    % \vspace{-1mm}
% \end{figure}

%-------------------------------------------------------------------------
\section{Conclusion}
Inspired by the philosophy of meta-learning, we propose a novel architecture named Meta Gradient Adversarial Attack to improve the transferability of adversarial examples. By iteratively simulating the scenarios of white-box and black-box attacks in the process of generating adversarial examples, the gap of gradient directions between black-box and white-box models can be reduced. Our architecture can be combined with any existing gradient-based attack method in a plug-and-play mode. Extensive experiments demonstrate that the adversarial examples generated by our architecture achieve better transferability. In future work, we will explore how to further improve the time efficiency of generating adversarial examples and analyze the reason why our method shows significant improvement under the targeted attack setting.

\section{Acknowledgment}
This work is partially supported by National Key R\&D Program of China (No. 2017YFA0700800), Natural Science Foundation of China (Nos. 61806188, 61976219), and Shanghai Municipal Science and Technology Major Program (No. 2017SHZDZX01).

%-------------------------------------------------------------------------
\clearpage
{\small
\bibliographystyle{ieee_fullname}
\bibliography{egbib}
}

\clearpage
\begin{appendices}

\section{Models}
\label{sec:models}
We list all the models used in our experiments here again for more friendly reading. And more details of these models are provided.

Our architecture utilizes a total of 10 white-box models to generate adversarial examples. In each iteration, multiple models are randomly selected to compose a meta-task. Under the scenario of black-box attack, we evaluate 6 and 7 models on ImageNet and CIFAR10, respectively. All the models used in white-box and black-box settings are shown in \tabref{tab:models}. For ImageNet, Ens3\_Inceptionv3, Ens4\_Inceptionv3, Ens\_InceptionResNetv2 and Adv\_Inceptionv3 are defense models adversarially trained with ensemble adversarial training~\cite{tramer2017ensemble}\footnote{\url{https://drive.google.com/drive/folders/10cFNVEhLpCatwECA6SPB-2g0q5zZyfaw}}. R\&P~\cite{xie2017mitigating}\footnote{\url{https://github.com/cihangxie/NIPS2017_adv_challenge_defense}}, NIPS-r3~\footnote{\url{https://github.com/anlthms/nips-2017/tree/master/mmd}} and CERTIFY~\cite{cohen2019certified} are also defense models. Other models are normally trained on ImageNet and publicly available\footnote{\url{https://github.com/tensorflow/models/tree/master/research/slim}}. For CIFAR10, Adv\_ResNet-18, Adv\_DenseNet-121 and Adv\_GoogLeNet are adversarially trained with FGSM, and Adv\_ResNet-18\_ll is adversarially trained with LeastLikely~\cite{kurakin2016adversarial}. $k$-WTA~\cite{xiao2019enhancing}, GBZ~\cite{li2019generative} and ADP~\cite{pang2019improving} are also defense models. The rest models are normally trained on CIFAR10.
\begin{table}[!hbtp]
   \begin{center}
      \caption{Models on CIFAR10 and ImageNet. The first 10 models are treated as white-box models and the rest models are black-box models in our experimental settings.}
      \label{tab:models}
      \vspace{-2mm}
      \resizebox{\columnwidth}{!}{
         \begin{tabular}{c|c|c|c}
            \hline 
            ~ & No. & ImageNet & CIFAR10 \\
            \hline
            \multirow{10}*{\begin{tabular}{c}white-\\box\\models\end{tabular}} & 1 & Inceptionv3~\cite{szegedy2016rethinking} & ResNet-18~\cite{he2016deep}\\ \cline{2-4}
            ~ & 2 & Inceptionv4~\cite{szegedy2016inception} & ResNetv2-18~\cite{he2016identity}\\ \cline{2-4}
            ~ & 3 & InceptionResNetv2~\cite{szegedy2016inception} & GoogLeNet~\cite{szegedy2015going}\\ \cline{2-4}
            ~ & 4 & ResNetv2-152~\cite{he2016identity} & ResNeXt-29~\cite{xie2017aggregated}\\ \cline{2-4}
            ~ & 5 & Ens3\_Inceptionv3~\cite{tramer2017ensemble} & SENet-18~\cite{hu2018squeeze}\\ \cline{2-4}
            ~ & 6 & Ens4\_Inceptionv3~\cite{tramer2017ensemble} & RegNetX-200mf~\cite{radosavovic2020designing} \\ \cline{2-4}
            ~ & 7 & Ens\_InceptionResNetv2~\cite{tramer2017ensemble} & DLA~\cite{yu2018deep}\\ \cline{2-4}
            ~ & 8 & ResNetv2-101~\cite{he2016identity} & Shake-ResNet-26\_2x64d~\cite{gastaldi2017shake}\\ \cline{2-4}
            ~ & 9 & MobileNetv2\_1.0~\cite{sandler2018mobilenetv2} & Adv\_ResNet-18\\ \cline{2-4}
            ~ & 10 & PNasNet~\cite{liu2018progressive} & Adv\_DenseNet-121\\ 
            \hline
            \multirow{7}*{\begin{tabular}{c}black-\\box\\models\end{tabular}} & 11 & Adv\_Inceptionv3~\cite{tramer2017ensemble} & PyramidNet-164~\cite{han2017deep}\\ \cline{2-4}
            ~ & 12 & NasNet\_mobile~\cite{zoph2018learning} & CbamResNeXt~\cite{woo2018cbam} \\ \cline{2-4}
            ~ & 13 & MobileNetv2\_1.4~\cite{sandler2018mobilenetv2} & Adv\_GoogLeNet \\ \cline{2-4}
            ~ & 14 & R\&P~\cite{xie2017mitigating} & Adv\_ResNet-18\_ll \\ \cline{2-4}
            ~ & 15 & NIPS-r3 & $k$-WTA~\cite{xiao2019enhancing} \\ \cline{2-4}
            ~ & 16 & CERTIFY~\cite{cohen2019certified} & GBZ~\cite{li2019generative} \\ \cline{2-4}
            ~ & 17 & ~ & ADP~\cite{pang2019improving} \\
            \hline
         \end{tabular}
      }
   \end{center}
   \vspace{-2mm}
\end{table}

\section{Impact of Hyperparameters}
\label{sec:hyperparameters}
In this section, we analyze the influence of the hyperparameters: the number of models $n$ selected for the ensemble-based attacks during the meta-train step and the number of tasks $T$ sampled during the entire generation. 

\textbf{The number of sampled tasks $T$.}
\begin{table*}[!hbtp]
   \begin{center}
      \caption{The attack success rates of the adversarial examples generated under \textbf{different sampled tasks $T$} against the white-box and black-box models on ImageNet. The number of iterations $K$ in meta-train step is 2. The number of ensembled models $n$ in meta-train step is 5. The index of models in the table is the same as \tabref{tab:models}.}
      \vspace{-2mm}
      \label{tab:task}
      \resizebox{\textwidth}{!}{
         \begin{tabular}{c|c|c|c|c|c|c|c|c|c|c|c|c|c|c|c|c}
            \hline 
            \multirow{2}*{$T$} & \multicolumn{10}{c|}{white-box models} & \multicolumn{6}{|c}{black-box models} \\ \cline{2-17}
            & 1 & 2 & 3 & 4 & 5 & 6 & 7 & 8 & 9 & 10 & 11 & 12 & 13 & 14 & 15 & 16 \\ \hline
            10 & 98.7 & 98.6 & 97.9 & 97.8 & 98.4 & 98.0 & 96.4 & 97.0 & 97.7 & 98.2     & 94.5 & 96.9 & 96.7 & 95.6 & 95.8 & 66.8  \\
            20 & 99.1 & 99.2 & 98.5 & 98.4 & 99.1 & 98.6 & 97.7 & 98.3 & 98.5 & 98.6     & 96.1 & 97.9 & 97.9 & 96.8 & 97.0 & 68.7  \\
            30 & 99.3 & 99.1 & 98.7 & 98.3 & 99.1 & 99.1 & 98.0 & 98.4 & 99.0 & 99.0     & 97.1 & 98.2 & 98.2 & 97.1 & 97.0 & 69.0  \\
            40 & 99.5 & 99.5 & 98.9 & 98.5 & 99.2 & 99.0 & 98.2 & 98.6 & 99.0 & 98.9     & 97.4 & 98.2 & 98.3 & 97.7 & 97.6 & 70.2  \\
            60 & 99.5 & 99.5 & 99.2 & 98.8 & 99.2 & 99.0 & 98.3 & \textbf{98.9} & 98.9 & 99.3     & 97.8 & 98.5 & 98.5 & 98.0 & \textbf{98.3} & 71.1 \\
            80 & 99.5 & 99.5 & \textbf{99.3} & \textbf{98.9} & 99.3 & 99.1 & 98.2 & 98.8 & 99.1 & 99.3     & 97.9 & 98.4 & \textbf{98.7} & 97.9 & 97.9 & 71.3  \\
            120 & \textbf{99.7} & \textbf{99.8} & \textbf{99.3} & 98.8 & \textbf{99.4} & \textbf{99.3} & \textbf{98.5} & \textbf{98.9} & \textbf{99.2} & \textbf{99.4}     & \textbf{98.2} & \textbf{98.7} & 98.6 & \textbf{98.1} & \textbf{98.3} & \textbf{71.6}  \\
            \hline
         \end{tabular}
      }
   \end{center}
   \vspace{-2mm}
\end{table*}
Our architecture simulates the process of white-box and black-box attacks iteratively by sampling different model combinations as a meta-task, and the number of sampled tasks $T$ may influence the attack success rate of the generated adversarial examples against the defenses. We compare the attack effects of the generated adversarial examples under white-box and black-box settings when $T$ ranges from 10 to 120 in \tabref{tab:task}. It can be seen that the more sampled tasks are taken, the higher attack success rate can be achieved, especially for black-box settings. The reason behind it lies in that more scenarios of white-box and black-box attacks are simulated by sampling more tasks and the gap of white-box and black-box attacks are gradually narrowed, leading to better transferability of adversarial examples against black-box models. However, on the other hand, increasing the number of sampled tasks also increases the time generating adversarial examples. Considering a trade-off of both efficiency and effectiveness, the value of $T$ is recommended to be 40.

\textbf{The number of ensembled models $n$ in meta-train step.}
\begin{table*}[!hbtp]
   \begin{center}
      \caption{The attack success rates of the adversarial examples generated under \textbf{different number of ensembled models $n$} in meta-train step against the white-box and black-box models on ImageNet. The number of sampled tasks $T$ is 40. The number of iterations $K$ in meta-train step is 5.} %The index of models in the table is the same as \tabref{tab:model}.}
      \vspace{-2mm}
      \label{tab:ens}
      \resizebox{\textwidth}{!}{
         \begin{tabular}{c|c|c|c|c|c|c|c|c|c|c|c|c|c|c|c|c}
            \hline 
            \multirow{2}*{$n$} & \multicolumn{10}{c|}{white-box models} & \multicolumn{6}{|c}{black-box models} \\ \cline{2-17}
            & 1 & 2 & 3 & 4 & 5 & 6 & 7 & 8 & 9 & 10 & 11 & 12 & 13 & 14 & 15 & 16 \\ \hline
            3 & \textbf{99.9} & 99.6 & 99.7 & 99.1 & 99.5 & 99.6 & 99.0 & 99.0 & 99.4 & 99.4      & 98.6 & 98.8 & 98.8 & 98.6 & 98.6 & 70.3 \\
            4 & 99.5 & 99.9 & \textbf{99.9} & 99.4 & 99.7 & 99.5 & 99.1 & 99.4 & 99.3 & 99.6      & 98.7 & 99.1 & 98.9 & 99.0 & 98.8 & 70.6 \\
            5 & \textbf{99.9} & \textbf{100} & 99.7 & 99.5 & 99.8 & 99.7 & 98.9 & 99.5 & \textbf{99.5} & 99.7    & 98.6 & \textbf{99.3} & 99.1 & 98.7 & 98.6 & 71.3 \\
            6 & 99.8 & 99.9 & 99.6 & 99.5 & \textbf{99.9} & \textbf{99.8} & 99.1 & \textbf{99.6} & 99.4 & 99.7   & 98.7 & 99.1 & 99.2 & 98.7 & 98.8 & 71.6 \\
            7 & \textbf{99.9} & 99.9 & 99.7 & 99.5 & 99.7 & \textbf{99.8} & 99.1 & 99.4 & 99.4 & 99.6   & \textbf{98.9} & \textbf{99.3} & 99.2 & 98.7 & 98.8 & \textbf{72.2} \\
            8 & \textbf{99.9} & \textbf{100} & 99.8 & \textbf{99.7} & \textbf{99.9} & \textbf{99.8} & 99.1 & \textbf{99.6} & 99.4 & 99.7    & 98.8 & \textbf{99.3} & \textbf{99.3} & 99.0 & \textbf{99.0} & 71.9 \\
            9 & \textbf{99.9} & \textbf{100} & 99.8 & \textbf{99.7} & \textbf{99.9} & \textbf{99.8} & \textbf{99.3} & 99.5 & 99.4 & \textbf{99.8}     & 98.6 & \textbf{99.3} & 99.2 & \textbf{99.1} & \textbf{99.0} & 72.0 \\
            \hline
         \end{tabular}
      }
   \end{center}
   \vspace{-2mm}
\end{table*}
In the meta-train step, we use an ensemble of multiple models to calculate the gradients and update the adversarial examples. We compare the attack success rates against the white-box and black-box models with an ensemble of different numbers of models $n$ during the update in \tabref{tab:ens}. It can be seen that, when the number of ensembled models increases, the success rates against the white-box and black-box attacks become higher and higher. But when $n$ is greater than 5, the increase in attack success rates is not obvious, which shows that our architecture is not sensitive to the hyperparameter $n$ to a certain extent. Considering that the more ensembled models in each iteration, the higher the computational complexity is needed. Therefore, it is a suitable choice to set the number of ensembled models to be 5.

\section{More Results of the Untargeted Attack}
\label{sec:more}
\begin{table*}[!hbtp]
   \begin{center}
      \caption{The attack success rates of the adversarial examples from our proposed Meta Gradient Adversarial Attack and some state-of-the-art methods against the white-box and black-box models on \textbf{ImageNet} under \textbf{untargeted attack} setting. The number of ensembled models $n$ in meta-train step is 5.  The number of iterations $K$ in meta-train step is 8.} %The index of models in the table is the same as \tabref{tab:model}.}
      \vspace{-4mm}
      \label{tab:sota_imagenets}
      \begin{subtable}[t]{\textwidth}
         \caption{The number of sampled tasks $T$ and the number of iteration in baseline methods are all 10.}
         \vspace{-2mm}
         \resizebox{\textwidth}{!}{
            \begin{tabular}{c|c|c|c|c|c|c|c|c|c|c|c|c|c|c|c|c|c}
               \hline 
               \multirow{2}*{Method} & \multicolumn{10}{c|}{white-box models} & \multicolumn{6}{|c|}{black-box models} & Time \\ \cline{2-17}
               & 1 & 2 & 3 & 4 & 5 & 6 & 7 & 8 & 9 & 10 & 11 & 12 & 13 & 14 & 15 & 16 & ($s$/img)\\ \hline
               SI-NI~\cite{lin2019nesterov}  & 99.6 &96.5 &96.1 &94.4 &98.8 &99.1 &92.4 &93.9 &99.0 &97.2 &43.0 &88.3 &90.5 &49.3 &53.1 &38.1 &22.76 \\ \hline
               MIM~\cite{dong2018boosting}   & 99.4 &99.5 &99.2 &98.2 &99.5 &99.8 &99.1 &98.6 &98.7 &97.7 &44.1 &91.5 &93.9 &66.0 &70.3 &34.9 &6.48  \\
               MGAA w/ MIM                   & \textbf{100} &\textbf{100} &\textbf{100} &\textbf{99.8} &\textbf{100} &\textbf{100} &\textbf{100} &\textbf{99.5} &\textbf{99.8} &\textbf{99.3} &\textbf{46.5} &\textbf{95.3} &\textbf{97.0} &\textbf{68.2} &\textbf{73.9} &\textbf{37.4} &22.54 \\ \hline
               DIM~\cite{xie2019improving}   & 99.5 &99.7 &99.4 &98.6 &99.5 &99.6 &98.5 &98.5 &98.8 &98.8 &78.5 &98.1 &98.3 &95.4 &87.7 &44.7 &8.23 \\
               MGAA w/ DIM                   & \textbf{100} &\textbf{100} &\textbf{100} &\textbf{99.6} &\textbf{99.9} &\textbf{99.8} &\textbf{98.9} &\textbf{99.6} &\textbf{99.7} &\textbf{99.6} &\textbf{79.9} &\textbf{99.3} &\textbf{99.4} &\textbf{96.5} &\textbf{97.3} &\textbf{48.7} &22.96 \\ \hline
               TI-DIM~\cite{dong2019evading} & 98.5 &98.5 &97.3 &97.2 &98.0 &97.7 &95.8 &97.1 &97.1 &97.7 &93.3 &96.3 &95.7 &95.1 &94.9 &67.8 &7.09 \\ 
               MGAA w/ TI-DIM                & \textbf{99.8} &\textbf{99.9} &\textbf{99.7} &\textbf{99.4} &\textbf{99.7} &\textbf{99.5} &\textbf{98.9} &\textbf{99.3} &\textbf{99.3} &\textbf{99.3} &\textbf{98.4} &\textbf{99.0} &\textbf{99.0} &\textbf{98.6} &\textbf{98.7} &\textbf{70.5} &21.33 \\ \hline
            \end{tabular}
         }
      \end{subtable}
      \begin{subtable}[t]{\textwidth}
         \caption{The number of sampled tasks $T$ and the number of iteration in baseline methods are all 40.}
         \vspace{-2mm}
         \resizebox{\textwidth}{!}{
            \begin{tabular}{c|c|c|c|c|c|c|c|c|c|c|c|c|c|c|c|c|c}
               \hline 
               \multirow{2}*{Method} & \multicolumn{10}{c|}{white-box models} & \multicolumn{6}{|c|}{black-box models} & Time \\ \cline{2-17}
               & 1 & 2 & 3 & 4 & 5 & 6 & 7 & 8 & 9 & 10 & 11 & 12 & 13 & 14 & 15 & 16 & ($s$/img)\\ \hline
               SI-NI~\cite{lin2019nesterov}  & 99.7 &97.5 &97.4 &96.3 &98.8 &98.6 &90.8 &95.6 &99.7 &98.2 &48.2 &90.8 &92.9 &50.6 &58.5 &38.9 &68.29 \\ \hline
               MIM~\cite{dong2018boosting}   & 99.6 &99.7 &99.4 &98.7 &99.8 &99.8 &99.5 &99.0 &99.1 &98.2 &44.4 &92.6 &94.1 &65.4 &72.2 &34.4 &17.51 \\
               MGAA w/ MIM                   & \textbf{100} &\textbf{100} &\textbf{100} &\textbf{99.9} &\textbf{100} &\textbf{100} &\textbf{100} &\textbf{99.9} &\textbf{99.9} &\textbf{99.9} &\textbf{52.0} &\textbf{96.0} &\textbf{96.9} &\textbf{67.1} &\textbf{74.9} &\textbf{37.0} &71.24 \\ \hline
               DIM~\cite{xie2019improving}   & 99.4 &99.8 &99.5 &98.6 &99.4 &99.5 &98.5 &98.6 &98.9 &98.8 &79.4 &98.0 &98.3 &95.0 &95.3 &44.8 &22.22 \\
               MGAA w/ DIM                   & \textbf{100} &\textbf{100} &\textbf{100} &\textbf{99.9} &\textbf{100} &\textbf{100} &\textbf{99.9} &\textbf{99.9} &\textbf{100} &\textbf{99.9} &\textbf{88.0} &\textbf{99.9} &\textbf{99.8} &\textbf{98.9} &\textbf{98.9} &\textbf{49.3} &69.26 \\ \hline
               TI-DIM~\cite{dong2019evading} & 98.9 &99.1 &98.2 &98.3 &98.9 &98.6 &97.3 &98.0 &98.1 &98.3 &96.3 &97.5 &97.5 &96.7 &96.8 &67.8 &19.13 \\ 
               MGAA w/ TI-DIM                & \textbf{100} &\textbf{100} &\textbf{99.9} &\textbf{99.8} &\textbf{99.9} &\textbf{99.8} &\textbf{99.4} &\textbf{99.8} &\textbf{99.6} &\textbf{100} &\textbf{99.1} &\textbf{99.4} &\textbf{99.5} &\textbf{99.4} &\textbf{99.0} &\textbf{71.6} &67.28 \\ \hline
            \end{tabular}
         }
      \end{subtable}
   \end{center}
   \vspace{-2mm}
\end{table*}

Except the experiments of the untargeted attack with $T$ being 40 illustrated in the manuscript, we also conduct the experiments with $T$ being 10, which is the common setting in MIM~\cite{dong2018boosting}, DIM~\cite{xie2019improving} and TIM~\cite{dong2019evading}. As shown in \tabref{tab:sota_imagenets}, our proposed MGAA also outperforms the state-of-the-art methods under the setting of $T$ being 10. Moreover, when comparing the results of our method under the setting of $T$ being 10 with the baseline methods under the setting of $T$ being 40, we can see that the time cost is nearly equal, but our method still achieves higher attack success rates under both white-box and black-box settings.

\begin{figure*}[htbp]
   \centering
   \begin{subfigure}[t]{0.31\textwidth}
      \centering
      \includegraphics[width=\textwidth]{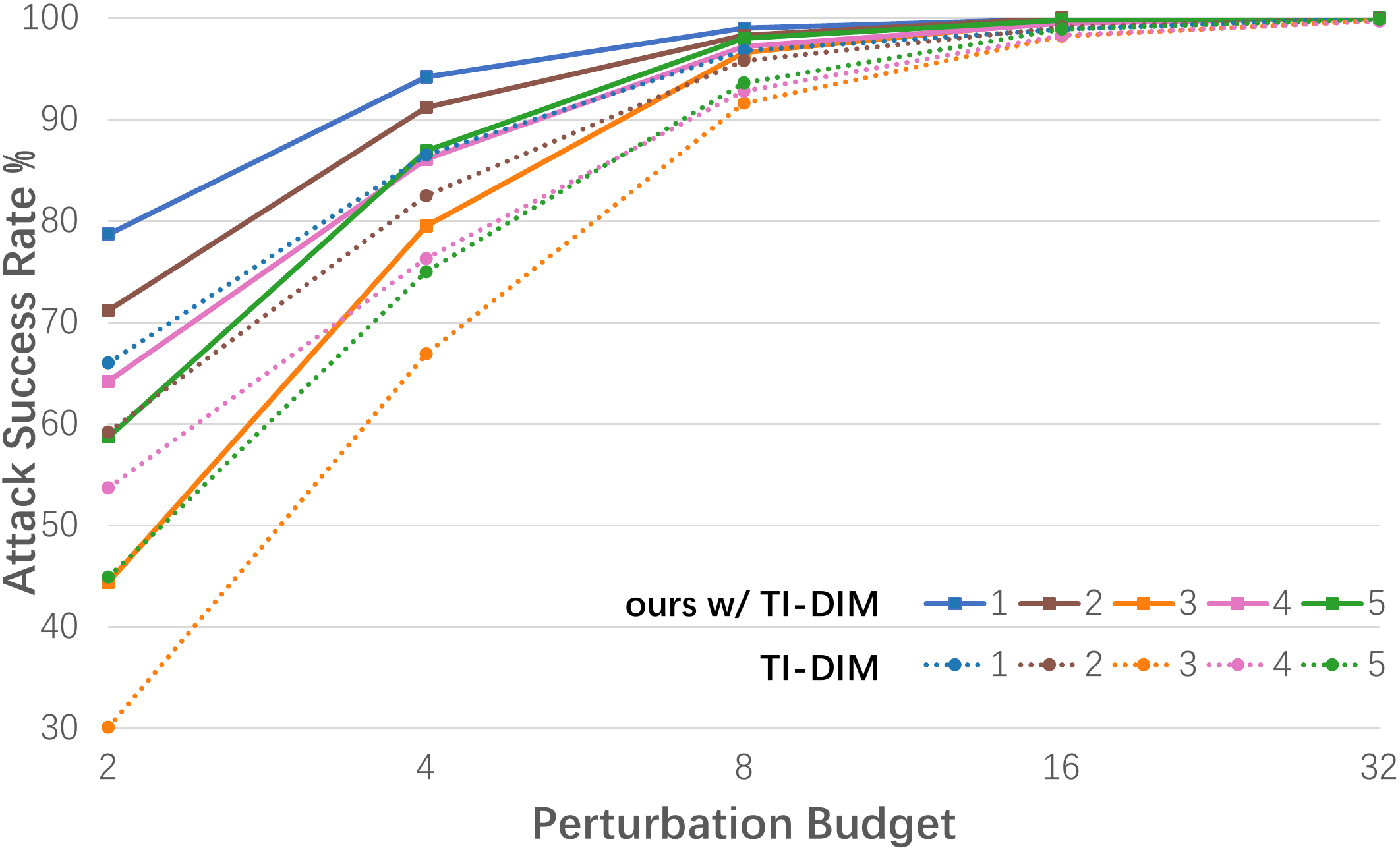}
      \end{subfigure}
      \quad
      \begin{subfigure}[t]{0.31\textwidth}
      \centering
      \includegraphics[width=\textwidth]{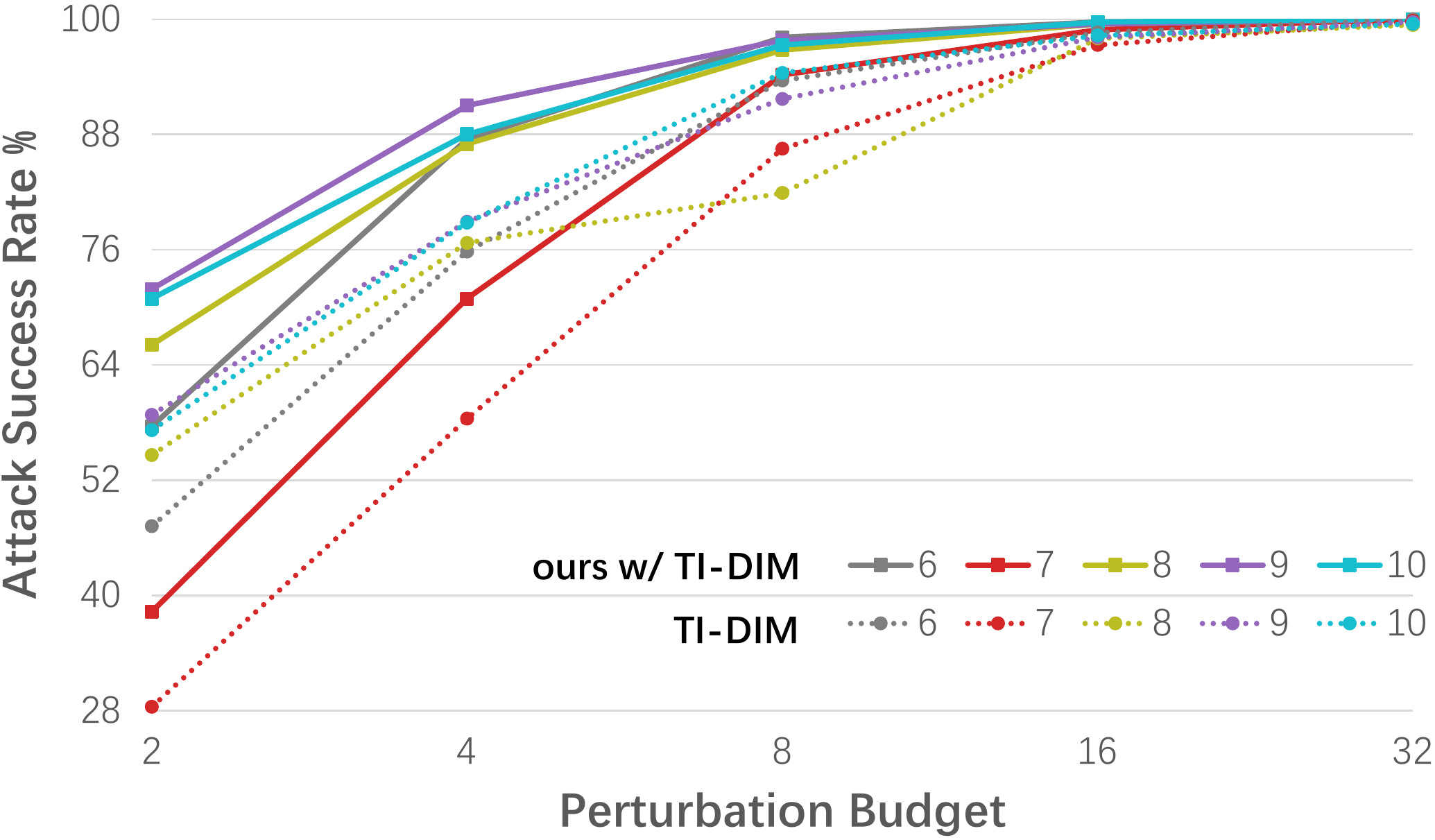}
   \end{subfigure}
   \quad
   \begin{subfigure}[t]{0.31\textwidth}
      \centering
      \includegraphics[width=\textwidth]{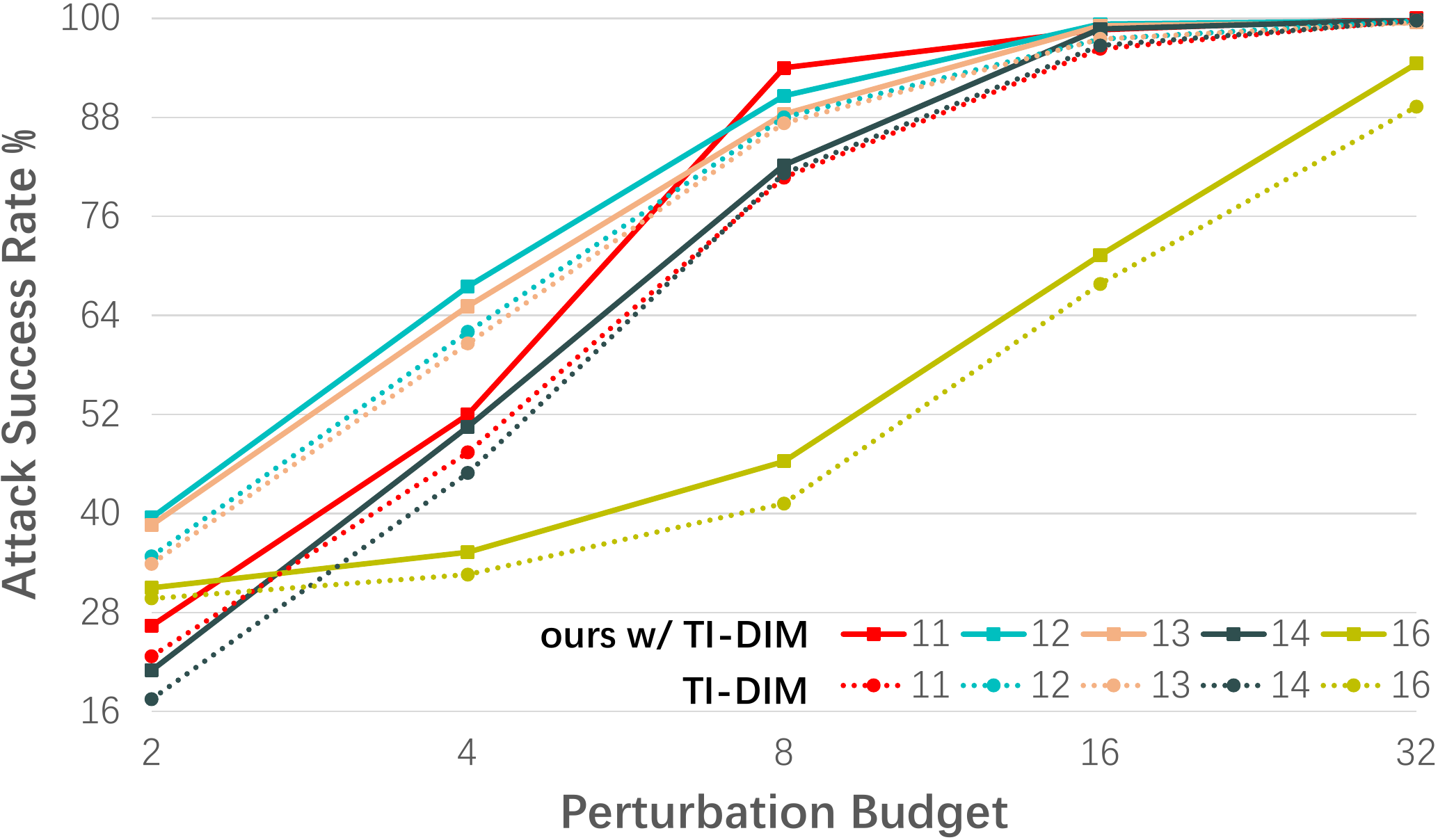}
   \end{subfigure}
   \vspace{-2mm}
   \caption{The attack success rates vs. perturbation budget curve on ImageNet. The dotted lines are the results of TI-DIM, and the solid lines are the results of our method. The number in the legend means the index of models in \tabref{tab:models}.}
   \label{fig:curve}
\end{figure*}

\begin{table*}[!htbp]
   \begin{center}
      \caption{The ablation study of different parts in our proposed MGAA architecture.}
      \vspace{-2mm}
      \label{tab:ablation}
      \resizebox{\textwidth}{!}{
         \begin{tabular}{c|c|c|c|c|c|c|c|c|c|c|c|c|c|c|c|c}
            \hline
            \multirow{2}*{Setting} & \multicolumn{10}{c|}{white-box models} & \multicolumn{6}{|c}{black-box models} \\ \cline{2-17}
            & 1 & 2 & 3 & 4 & 5 & 6 & 7 & 8 & 9 & 10 & 11 & 12 & 13 & 14 & 15 & 16 \\ \hline
            MGAA w/o meta-train & 95.7 &93.7 &92.4 &96.0 &94.3 &93.1 &89.3 &96.3 &97.9 &96.2 &81.5 &89.4 &89.3 &81.8 &82.6 &52.5 \\
            MGAA w/o meta-test & 98.9 &99.1 &98.2 &98.3 &98.9 &98.6 &97.3 &98.0 &98.1 &98.3 &96.3 &97.5 &97.5 &96.7 &96.8 &67.8 \\
            MGAA & \textbf{100} &\textbf{100} &\textbf{99.9} &\textbf{99.8} &\textbf{99.9} &\textbf{99.8} &\textbf{99.4} &\textbf{99.8} &\textbf{99.6} &\textbf{100} &\textbf{99.1} &\textbf{99.4} &\textbf{99.5} &\textbf{99.4} &\textbf{99.0} &\textbf{71.4} \\
            \hline
         \end{tabular}
      }
   \end{center}
   \vspace{-2mm}
   Note: MGAA w/o meta-test is actually the same as existing method like TI-DIM~\cite{dong2019evading}.
   \vspace{-2mm}
\end{table*}

\begin{table}[!htbp]
   \begin{center}
      \caption{The cosine similarity between the generated adversarial perturbations on ten \textbf{white-box models} and the gradient directions directly computed on three \textbf{black-box models}.}
      \vspace{-3mm}
      \label{tab:cosine}
      \resizebox{\columnwidth}{!}{
         \begin{tabular}{c|c|c|c}
            \hline
            Method & Adv\_Inceptionv3 & NasNet & MobileNetv2\_1.4 \\
            \hline
            TI-DIM & -0.104 & -0.070 & -0.084 \\
            MGAA w/ TI-DIM & 0.113 & 0.071 & 0.083 \\
            \hline
         \end{tabular}
      }
   \end{center}
   % \vspace{-6mm}
\end{table}

\section{Attack under Various Perturbation Budgets}
\label{sec:budget}

We conduct experiments of the attack under various perturbation budgets. The curve of attack success rate vs. perturbation budgets is shown in \figref{fig:curve}. The curve with dotted lines are the results of TI-DIM, and the curve with solid lines are the results of our method. We can clearly see that our MGAA consistently achieves higher attack success rates in both white-box and black-box attacks under various perturbation budgets.

\section{Ablation Study}
\label{sec:ablation}
We conduct an ablation study to demonstrate the effectiveness of each part in our proposed MGAA architecture. The version of MGAA without meta-test is actually the existing methods like TI-DIM~\cite{dong2019evading}, \ie, using an ensemble of multiple models to update the adversarial examples in each step of the iteration. The version of MGAA without meta-train is to use only one randomly selected model to update the adversarial perturbation in each step. From \tabref{tab:ablation}, we can see that the meta-train step plays a more important role than the meta-test step. And the full version of our MGAA architecture achieves higher attack success rates than both meta-train only and meta-test only versions.

\section{The Cosine Similarity of the Gradients}
\label{sec:cosine}
We calculate the cosine similarity between the generated adversarial perturbations on ten \textbf{white-box models} and the gradient directions of three \textbf{black-box models}, \ie, Adv\_Inceptionv3, NasNet, and MobileNetv2\_1.4. The range of cosine similarity is -1 to 1, and the bigger value means the more similar direction. As shown in \tabref{tab:cosine}, the generated adversarial perturbations by our MGAA are closer to the gradient directions of both three black-box models consistently, which verifies the theoretical analysis in Sec. 3.3 in the paper that MGAA can narrow the gaps of gradient directions between the white-box and black-box models.

\section{Minimum Adversarial Noises}
\label{sec:min_noise}
\begin{table}[!htbp]
   \footnotesize
   \begin{center}
      \caption{The mean of minimum adversarial noises needed to fool images on ImageNet.}
      \label{tab:min}
         \begin{tabular}{c|c|c|c}
            \hline 
            Method & $L_{\infty}$ & $L_1$ & $L_2$ \\
            \hline
            TI-DIM & 7.4576 & 5.3112 & 9.8975 \\ 
            MGAA w/ TI-DIM & 4.2960 & 3.1203 & 5.8767 \\
            \hline
         \end{tabular}
   \end{center}
\end{table}

We conduct the experiment to see the minimum adversarial noise needed to fool each image. From \tabref{tab:min} we can see that the minimum adversarial noise needed in our MGAA is less than TI-DIM method under various metric evaluations.

\end{appendices}

\end{document}